\documentclass[conference]{IEEEtran}
\usepackage{times}

\usepackage[numbers]{natbib}
\usepackage{multicol}
\usepackage{amsmath}
\usepackage{amssymb}  
\usepackage{wrapfig}
\usepackage{mathptmx} 
\usepackage{adjustbox}
\usepackage{booktabs, makecell}
\usepackage{multirow}
\usepackage{times}
\usepackage{textcomp}
\usepackage{floatrow}
\DeclareMathAlphabet{\mathcal}{OMS}{cmsy}{m}{n}
\usepackage[normalem]{ulem}
\usepackage{adjustbox}

\newcommand{\norm}[1]{\left\lVert#1\right\rVert}
\DeclareMathOperator*{\argmax}{argmax}

\begin{document}

\title{Never Mind the Bounding Boxes,\\ Here's the SAND Filters}

\author{Author Names Omitted for Anonymous Review. Paper-ID [add your ID here]}
\author{Zhiqiang Sui, Zhefan Ye and Odest Chadwicke Jenkins}

\author{\authorblockN{Zhiqiang Sui, Zhefan Ye and Odest Chadwicke Jenkins}
\authorblockA{Department of Electrical Engineering and Computer Science, University of Michigan \\
Email: \{zsui, zhefanye, ocj\}@umich.edu}
}


%

\maketitle
\begin{abstract}
Perception is the main bottleneck to perform autonomous mobile manipulation tasks, especially in cluttered and unstructured environment. In this paper, we propose a novel two-stage paradigm that leverage both CNN object prior and generative sampling to perform object detection and $6D$ pose estimation. Our two-stage approach builds upon both CNN and generative sampling-based local search method to achieve sampling the network density, or SAND filter. We show the quantitative results that SAND effectively improve object detection result by reducing false positive and false negative recognitions, and further produces accurate pose estimation. We also conduct extensive categorical object sorting  experiments to show our method is able to produce accurate and reliable detections and object poses.
\end{abstract}

\IEEEpeerreviewmaketitle


\section{Introduction}
\begin{figure}[t]
\begin{center}
   \includegraphics[width=0.9\linewidth]{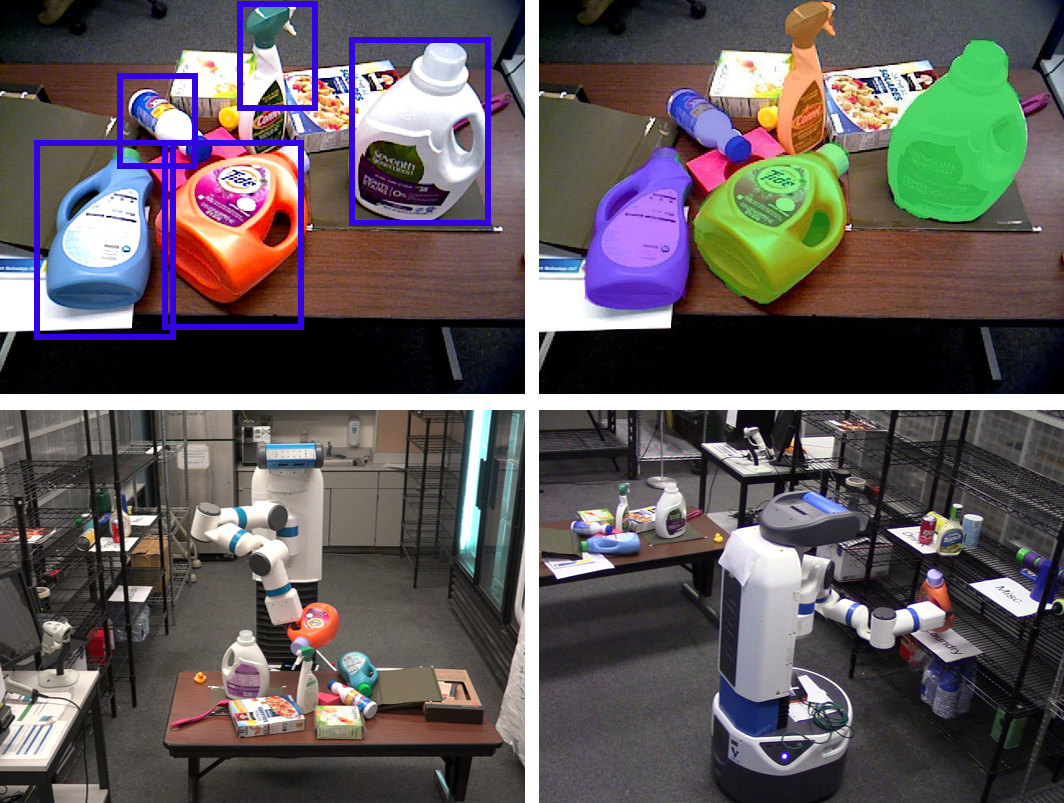}
\end{center}
   \caption{
    A robot perceiving and sorting objects from a cluttered tabletop. The goal is to move objects from the table to the shelf based on their categories. Our SAND filter enables the robot to detect objects and estimate their poses in a two-stage process.}
\label{fig:teaser}
\end{figure}

Robust and reliable operation of autonomous mobile manipulators remains an open challenge for robotics, where perception remains a critical bottleneck. Within the well-known sense-plan-act paradigm, truly autonomous robot manipulators need the ability to perceive the world, reason over manipulation actions afforded by objects towards a given goal, and carry out these actions in terms of physical motion. However, performing manipulation in unstructured and cluttered environments is particularly challenging due to many factors. Particularly, to execute a task with specific grasp points demands first recognizing object and estimating its precise pose. Figure~\ref{fig:teaser} illustrates such a task. The robot moves the objects from the table to the shelf based on their categories. 

With the advent of convolutional neural networks (CNN), many challenging perception tasks have been improved significantly, such as image classification~\cite{alexnet} and object detection~\cite{rcnn}. However, relying solely on CNN to detect objects and estimate their poses in a cluttered environment poses several issues. For instance, varying orientations and occlusions may greatly alter the appearance of objects, which may affects the performance of object detectors. Therefore, to avoid making hard decisions on the detection result, we present an alternative paradigm for object detection as well as pose estimation so that it can not only utilize the discriminative power given by deep neural networks but also maintain the versatility and robustness despite the everchanging environments.

Our two-stage approach builds upon both CNN and generative sampling-based local search method to achieve sampling the network density, or \emph{SAND filter}. The first stage of SAND filter attempts to detect objects using CNN and RGB images. However, unlike other popular object detectors, such as Faster RCNN~\cite{faster-rcnn}, we do not perform any filtering over the object bounding boxes, no matter their object confidence scores. Hence, the sampling method in the second stage could take full advantage of the probability density prior provided by CNN. Generative sampling method has been widely used in robot localization~\cite{dellaert1999monte} and object tracking~\cite{isard1996contour} due to its robustness. we, instead, employ such methods to perform local search over the hypothesis space in observed depth images so as to refine the object detection results as well as estimate object poses. Each sample, in our second stage, represents a possible object pose. The weight of each sample is bootstrapped by the prior given by the CNN and re-sampled based on its hypothesized states, from the rendering engine, and the observed state from the depth image. After iterations, the samples will essentially converge and the final state, which can best represent the observed scene.

\begin{figure*}[t]
\begin{center}
   \includegraphics[width=0.9\linewidth]{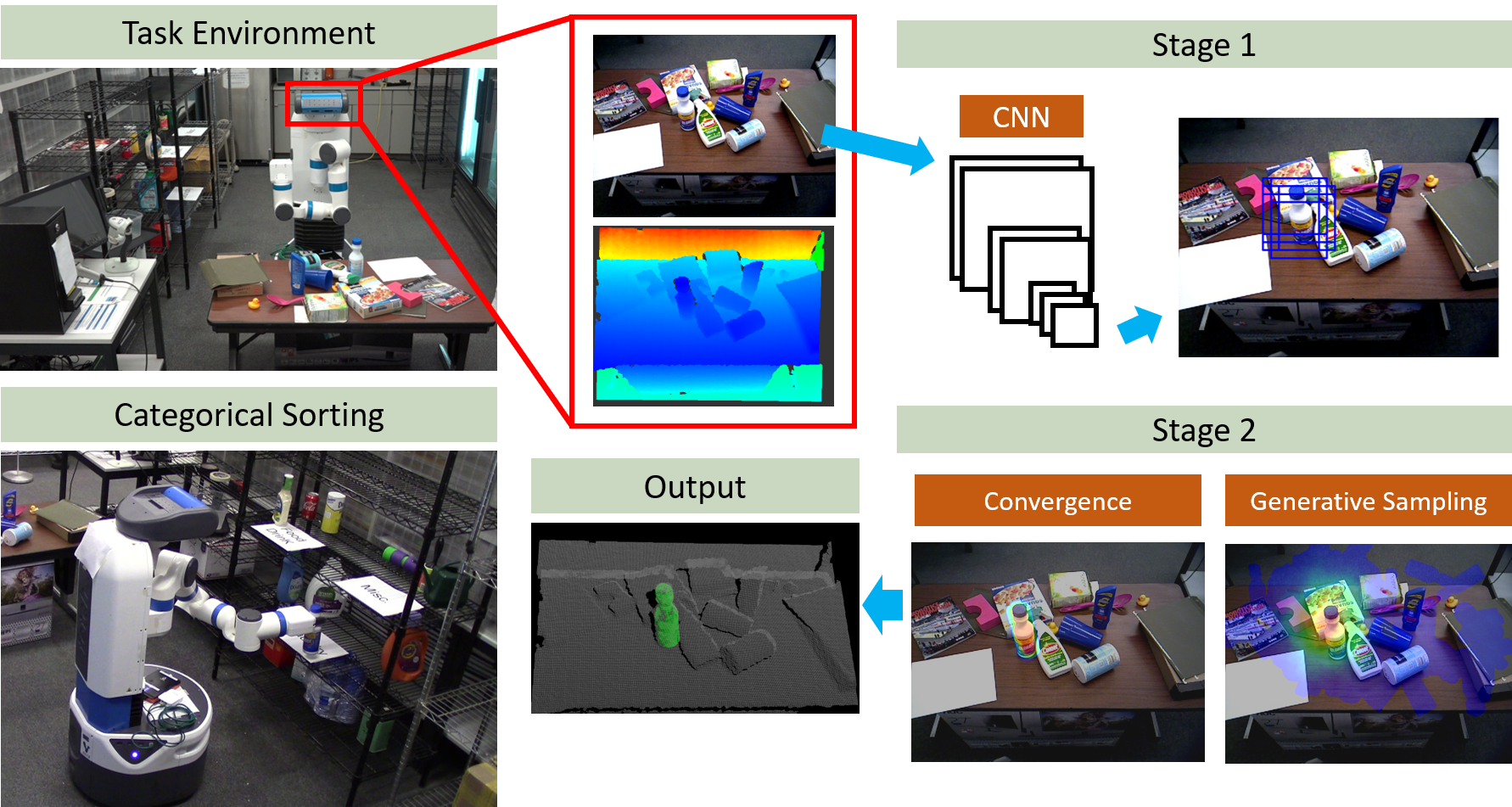}
\end{center}
   \caption{Overview of our approach. The robot operates in a cluttered environment and it captures an RGB-D image. Stage 1 takes in an RGB image and generate object bounding boxes with confidence scores. Stage 2 crops the corresponding depth image with the bounding boxes and perform bootstrap filtering to estimate the poses for all objects in the scene. Eventually, the robot performs object manipulation based on the estimated poses.}
\label{fig:overview}
\end{figure*}

In this paper, we demonstrate that the second stage of SAND filter can effectively improve object detection result from the first stage, and further produces accurate pose estimation. It is worth noting that our SAND filter does not limit to our own CNN implementation, and it can be adapted to other CNN-based object detectors with minor modification. Likewise, the sampling-based local search in the second stage can also be replaced.

\section{Related Work}
\subsection{Perception for Manipulation}
PR2 interactive manipulation~\cite{ciocarlie2014towards} segmented non-touching objects from a flat surface by clustering of surface normals. Collet et al. presented a discriminative approach, MOPED, to detect object and estimate object pose using iterative clustering-estimation (ICE) using multiple cameras~\cite{collet2011moped}.  Papazov et al.~\cite{papazov2012rigid} used a bottom-up approach of matching the 3D object geometries using RANSAC and retrieval by hashing methods. Narayanan et al. ~\cite{Narayanan-RSS-16} integrate A* global search with the heuristics from the neural networks to perform scene estimation assuming known identification of objects.

For manipulation in the cluttered environment, Ten Pas et al.~\cite{ten2016localizing} have shown success to detect the handle-like part of the object for grasp poses in cluttered environment and ~\cite{gualtieriPP17} et al. tried to perform the pick and place task in the deep reinforcement learning framework. Varley et al.~\cite{Varley2015IROS} developed a grasp pose generation system under partial view of the objects using deep learning.

\subsection{Object Detection}
Followed by the success of AlexNet~\cite{alexnet}, regions with convolutional neural network features, or R-CNN~\cite{rcnn}, introduced by Girshick et al., has become the dominating method for object detection. It first utilizes low/mid-level features to generate object proposal~\cite{ss},~\cite{edgebox}, and then uses CNN to extract feature within each proposal. Finally, a linear classifier, such as SVM~\cite{vapnik97nips}, is trained using those features for the classification task. \cite{sppnet} and~\cite{fast-rcnn} further optimized feature extraction process in R-CNN. Unified approaches, which integrate object proposal and classification, including our work, are inspired by R-CNN.

Long et al. propose FCN for semantic segmentation by replacing fully connected layers in traditional CNN with $1\times1$ convolutional layers~\cite{fcn}. FCN take images of arbitrary size and provide per-pixel classification label. However, FCN are not able to separate neighboring objects within the same category to obtain instance-level label; hence we cannot directly re-task FCN for object detection purpose. Nonetheless, most unified approaches are based on FCN to localize and classify objects using the same networks.

Recently, there has been a trend to utilize FCN to perform both object localization and classification~\cite{overfeat},~\cite{yolo},~\cite{faster-rcnn},~\cite{densebox}.

Sermanet et al. propose a integrated CNN framework for classification, localization and detection in a multiscale and sliding window fashion~\cite{overfeat}. All three tasks are learned simultaneously using a same shared network. Ren et al. expand the approach of~\cite{fast-rcnn} by taking the same networks for classification task and repurposing them for generating object proposals. Redmon et al. use a different approach by dividing the image into regions using a single network, and predicting bounding boxes and classification score for each region~\cite{yolo}.

Our approach is inspired by Faster R-CNN~\cite{faster-rcnn} and pose estimation by generative sampling methods~\cite{sui2015iros}, ~\cite{sui2017ijrr}, ~\cite{sui2017iros}. However, our approach proposes a two-stage framework to further address the challenge that detecting objects and estimating poses in a cluttered environment by enabling the generative sampling methods with the full potential of deep neural networks.

\begin{figure}[t]
    \centering
    \includegraphics[width=0.7\linewidth]{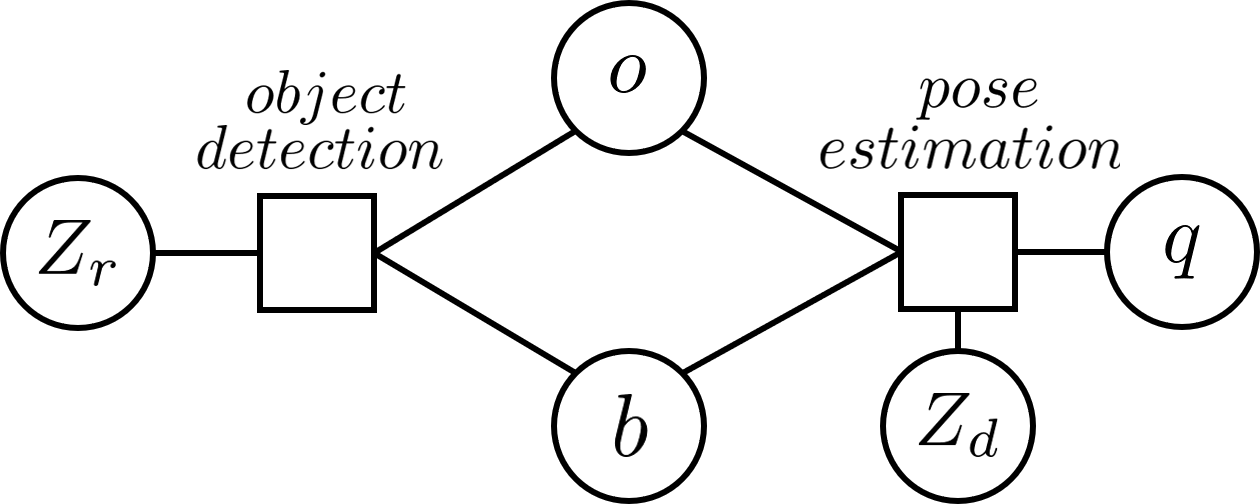}
    \caption{Factor graph representation of our two-stage approach.}
\label{fig:pgm}
\end{figure}

\section{Problem Formulation}

Given an RGB-D observation ($Z_r$, $Z_d$) from the robot sensor, our aim is to estimate the joint distribution $P(q, b, o, Z_r, Z_d)$, where $q$ is the six DoF object pose in the world frame, which comprises 3D spatial location and 3D orientation, $b$ is the object bounding box with a confidence score in the 2D image-space and $o$ is the object label with its corresponding 3D geometry model. Figure~\ref{fig:pgm} illustrates the formulation using factor graph for each object $o$ and can be represented as the following formulation:

\begingroup\makeatletter\def\f@size{9}\check@mathfonts
\begin{align}
&P(q, b, o, Z_r, Z_d) \label{eq:joint}\\
&= P(q \mid b, o, Z_r, Z_d)\ P(b, o, Z_r, Z_d) \label{eq:chain1} \\
&= P(q \mid b, o, Z_r, Z_d)\ P(b \mid o, Z_r, Z_d)\ P(o, Z_r, Z_d) \label{eq:chain2}\\
&= \underbrace{P(q \mid b, o, Z_d)}_\text{pose estimation} \underbrace{P(b \mid o, Z_r)}_\text{object detection} \underbrace{P(o, Z_r, Z_d)}_\text{observation} \label{condind}  \\
&\propto P(q \mid b, o, Z_d)\ P(b \mid o, Z_r) \label{eq:formulation}
\end{align}
\endgroup

Equation \ref{eq:chain1} and \ref{eq:chain2} are derived using \emph{chain rule statistics} and equation \ref{condind} represents the factoring of object detection, pose estimation and the observation prior. 
Here, we assume that pose estimation is \emph{conditionally independent} of RGB observation, $Z_r$, and object detection is \emph{conditionally independent} of depth observation, $Z_d$.

Ideally, we could use \emph{Markov chain Monte Carlo (MCMC)} \cite{Hastings_MCMC} to estimate the distribution of Equation~\ref{eq:joint}. However, the state space of the entire states are so large which makes it intractable to directly compute. End-to-end neural network method can also be also used to calculate the distribution. For instance, PoseCNN attempts to estimate object pose given RGB images only within a single CNN framework~\cite{xiang2017posecnn}. However, PoseCNN requires significant amount of data and human annotation in order to train the CNN. Our paradigm, on the other hand, is able to compensate the data deficiency by employing a generative sampling method in the second stage. SUM~\cite{sui2017iros} implements a simple combination of Equation~\ref{eq:joint} to enable sequential manipulation. However, in~\cite{sui2017iros}, data association is required to track the location of objects over time, which may lead to prolonged mis-detections if given malignant initial estimation. Furthermore, SUM may suffer from inevitable false detection, and hence, poor pose estimation, because a \emph{hard} filtering is performed after object proposal and detection stage.

\section{Inference Method}
We propose a two stage paradigm to compute object detection factor and pose estimation factor in two stages. Figure~\ref{fig:overview} illustrates the overview of our method. Our robot first has to estimate pose for each object given an RGBD image under cluttered environment. In the first stage, our CNN localize objects and give each bounding boxes a confidence score. Then, in the second stage, we perform generative sampling-based optimization to estimate the object pose given a depth image and object bounding boxes with scores. Once the pose of an object has been estimated, the robot picks up the object and places it on a shelf categorically. We hope that the heuristics from the first stage can better inform the generative sampling optimization in the second stage and the generative sampling can help check on the false detections from the first stage. 

Instead of directly computing Equation~\ref{eq:joint} because of the curse of dimensionality, we aim to maximize the joint probability by finding a pose $q*$ and can be defined as
\begin{align}
&\argmax_q P(q \mid b, o, Z_r, Z_d) P(b, o, Z_r, Z_d)
\end{align}
Thus, due to the nature of $\argmax$, our paradigm is limited to estimating one instance of an object class in the scene. However, we will extend our paradigm to accommodate multiple instances by estimating the joint probability in the future. 

\begin{figure}[t]
\begin{center}
   \includegraphics[width=0.8\linewidth]{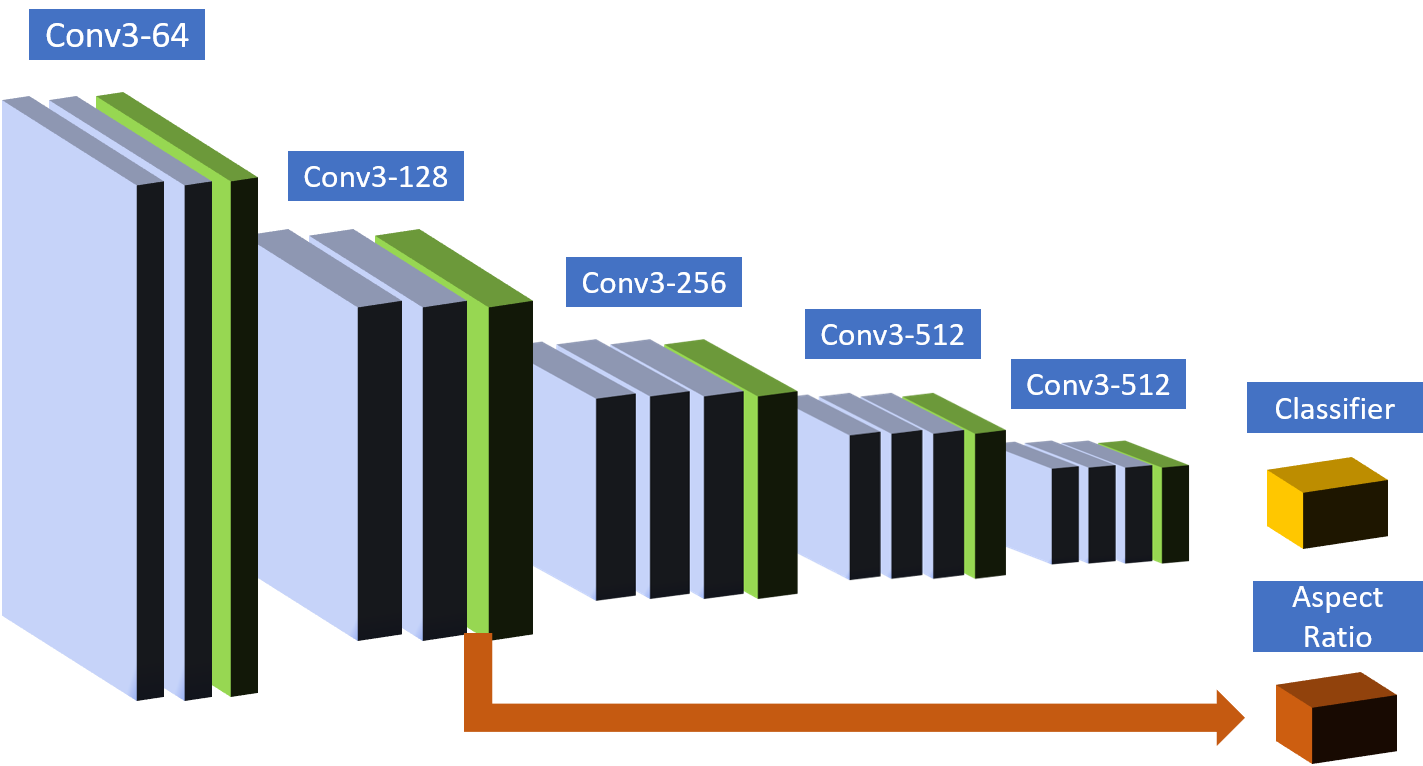}
\end{center}
   \caption{Network Architecture. Our CNN are based on VGG-16 network architecture, and have two branches, classifier and aspect ratio, to detect objects. The blue layers are convolutional layers and green layers are max pooling layers.}
\label{fig:network}
\end{figure}

\subsection{Object Detection}
\label{sec:s1}
The goal of our stage one method is to provide object bounding boxes with confidence scores given an object class $o$. To achieve this, we exploit the discriminative power of CNN. Inspired by region proposal networks (RPN) in~\cite{faster-rcnn}, our CNN serve as a \emph{proposal} method for the second stage. However, instead of only classifying objects as object and non-object, our networks are able to produce the object class labels. 

We choose VGG-16 networks~\cite{vgg} as our base networks. VGG-16 has $16$ weight layers coupled with ReLU layers and max pooling layers. To enable VGG-16 to perform object detection at each window location, we replace fully connected layers with $1\times1$ convolutional layers to construct fully convolutional networks (FCN), such as in~\cite{fcn}. Consequently, our networks ``convolve'' the input RGB image in a sliding window fashion.

Beyond the classification output, we would also like to predict the aspect ratio of the object bounding box. After \emph{conv2} layer, our CNN extend another branch to predict the aspect ratio of the bounding boxes, which is class agnostic. The detailed architecture is illustrated in Figure~\ref{fig:network}. Here, we choose not to adopt bounding box regression or \emph{anchors} method. Because of occlusions and varying poses of objects(Figure~\ref{fig:teaser}), it's challenging to estimate the exact 2D coordinates of objects with regression or predict aspect ratios based on statistics. Hence, our shape branch of CNN does not require class specific feature from later layers~\cite{zeiler2014visualizing}, such as \emph{conv5}, and simply intends to provide aspect ratio prior and leave the exact localization to the second stage.

The input to our networks is a pyramid of images with different scales. This is to enable the networks to capture objects of all sizes. The output, thus, is a \emph{pyramid} of heatmaps with another \emph{pyramid} of shape maps. Each pixel in the heatmap, with the corresponding pixel in the shape map, represents a bounding box in the input image. For each bounding box, there is a categorical distribution that represents possible outcomes of all object classes. Therefore, the bounding boxes, $b$, received by the next stage, is a list of 5-tuple that represents object 2D coordinates and a confidence score, $\{x_{min}, y_{min}, x_{max}, y_{max}, c\}$, given an object class $o$.

\subsection{Pose Estimation}
\label{sec:s2}
The purpose of the second stage is to estimate the object pose, $q$, and further refine the bounding box, $b$, based on the estimated pose, by performing generative sampling-based local search. Our local search is inspired by sampling methods, such as bootstrap filter~\cite{gordon1993novel}, which offer us robustness and versatility over the search space, which is critical in our context, since the result of the first stage may be imperfect and the manipulation task depends on the accuracy of the pose. Hence, we expect the second stage to improve object localization, or even correct false detections, based on the result of the first stage.


A collection of $M$ weighted samples, $\{q^{(i)},\ w^{(i)}\}_{i=1}^{M}$, to represent multiple hypotheses that indicate the states of object poses. Given an object class $o$, we have a corresponding object geometry model 
, and therefore can render a point cloud, $r$, using the z-buffer from a 3D graphics engine, given an object pose and camera view matrix 
. Essentially, these rendered point clouds are our collection of samples 
.

The initial samples are determined by the output of the first stage. Recall that in Section~\ref{sec:s1}, our CNN produce a density pyramid which is a list of bounding boxes with confidence scores, $\{x_{min}, y_{min}, x_{max}, y_{max}, c\}$. We perform the \emph{importance sampling} over the confidence scores $c$ to initialize our samples. The bounding box with a higher confidence score will get more samples to search over.

To evaluate a sample state, we first crop the depth image $Z_d$ with the corresponding bounding box, $\{x_{min}, y_{min}, x_{max}, y_{max}\}$ and then back-project it into a point cloud, $z^{(i)}$.
Note that $z^{(i)}$ can be different for different samples as they associate with different bounding boxes. We measure the ``similarity'' between the rendered point cloud and observed point cloud by counting how many points they match with each other. First, we define the \emph{inlier} function as the following, 
\begin{equation}
\begin{split}
  	\text{Inliers}(p, p^{\prime}) =  
  &\begin{cases}
    1, & \text{if $\norm{p - p^{\prime}}_{2} < \epsilon$}\\
    0, & \text{otherwise} \\
  \end{cases}
\end{split}
\end{equation}
\noindent where $p$ is a point in the observed point cloud $z^{(i)}$, and $p^{\prime}$ is a point in the rendered point cloud, $r$. If the Euclidean distance between an observed point and a rendered point is within a certain sensor resolution range, $\epsilon$, the total number of \emph{inliers} will increase by 1. The number of inliers is then defined as
\begin{equation}
N^{(i)} = \sum_{a, b \in z^{(i)}} \text{Inliers}(r(a, b), z^{(i)}(a, b))
\end{equation}
\noindent where $a$ and $b$ are 2D indices in the observed point cloud $z^{(i)}$.
Next, the weight $w^{(i)}$ of each hypothesis $q^{(i)}$ is defined as,
\begin{equation} 
W(q^{(i)}) = \alpha * \frac{N^{(i)}}{N_b} + \beta * \frac{N^{(i)}}{N_r} + \gamma * c \label{eq:weight}
\end{equation}
where $N_b$ is the number of points in the observed point cloud $z^{(i)}$, $N_r$ is the number of points in the rendered point cloud and $\alpha$, $\beta$ and $\gamma$ are coefficients. The first term in equation \ref{eq:weight} weighs how much the rendered point cloud match within the bounding box with the observed point cloud. However, since the bounding boxes from Section~\ref{sec:s1} are not perfect and usually truncate the objects, the second term is used to accommodate this by weighing how much the current hypothesis can explain itself not only in the bounding box but in the scene. We further blend in the object confidence score $c$ from the previous stage to balance between the two stages. 

To get the the optimum pose $q^{*}$, we follow the procedure of \emph{importance sampling} to assign a new weight to each sample. During the re-sampling process, each pose, $q^{(i)}$, would also be perturbed by a normal distribution in the space of six DoF. Once the average weight is above a threshold, $\tau$, we consider the local search is converged, and $q^{*}$, which is the sample with maximum weight, can best approximate the true object pose.

\section{Implementation}
We use PyTorch\footnote{http://pytorch.org/} for our CNN implementation. The aspect ratio branch of can predict $7$ aspect ratios. One training image contains only one object, and the size is $224\times224$. The aspect ratio of an object in the training image can be inferred from the width and height of the object. We ignore the aspect ratio labels of background images during training. The activation function for both branches of our CNN is \emph{softmax} since we consider predicting aspect ratio as a classification task as well. Thus, the loss function in training phase is \emph{cross entropy}.

Our second stage local search method relies on OpenGL graphics engine to render depth images, given a 3D geometry model and a camera view matrix. During the local search process, we allocate $625$ samples for each iteration and perform $200$ iterations in total. After the final iteration, we select the sample with highest weight and consider it to be the estimated object pose.

Followed by pose estimation, we further generate grasp poses of the object based on the method in~\cite{ten2016localizing}. To perform categorical sorting task as illustrated in Figure~\ref{fig:overview}, we use a Fetch robot to perform object grasping and placement. The Fetch robot is a mobile manipulator with seven DoF arm and a pan-tilt head equipped with an RGB-D sensor.

\section{Experiments}

\begin{figure}[t]
    \centering
    \includegraphics[width=0.9\linewidth]{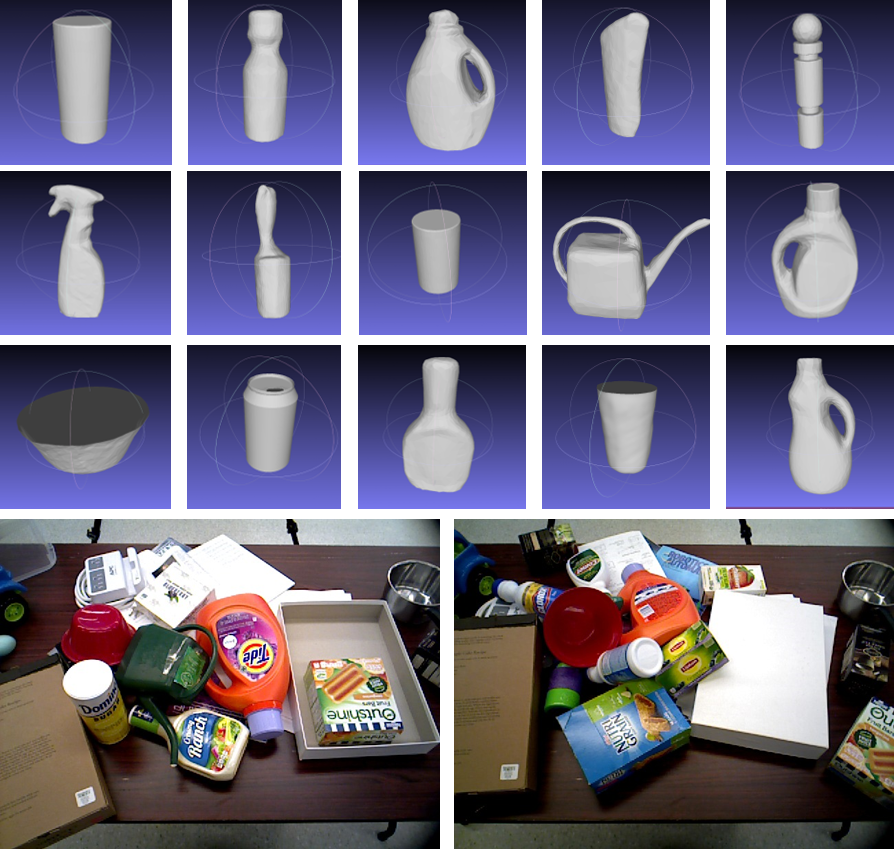}
    \caption{Tabletop object dataset. We have 15 objects total. For each scene, we randomly place a set of objects in the scene.}
\label{fig:dataset}
\end{figure}

\subsection{Dataset}
Our dataset contains $58$ images with $15$ object classes and are collected in a cluttered tabletop environment as shown in Figure~\ref{fig:dataset}. All objects are labelled with 2D bounding boxes and six DoF poses. We use $30$ images for training and parameter tuning and $28$ images for testing.

\subsection{Results}
We first present the result and analysis on pose estimation accuracy, followed by the robot categorical sorting tasks. 

\subsubsection{Pose Estimation}
In the estimation experiments, we evaluated the performance of SAND filter method on 28 test images in the dataset. We employ the evaluation metric from~\cite{hinterstoisser2012accv} to measure the mean of the pairwise point distance between the ground truth point cloud and the rendered point cloud from the estimated 6D pose. This metric can also account for objects with symmetric axis, which is useful in comparing household objects. 

\begin{figure}[t]
    \centering
    \includegraphics[width=\linewidth]{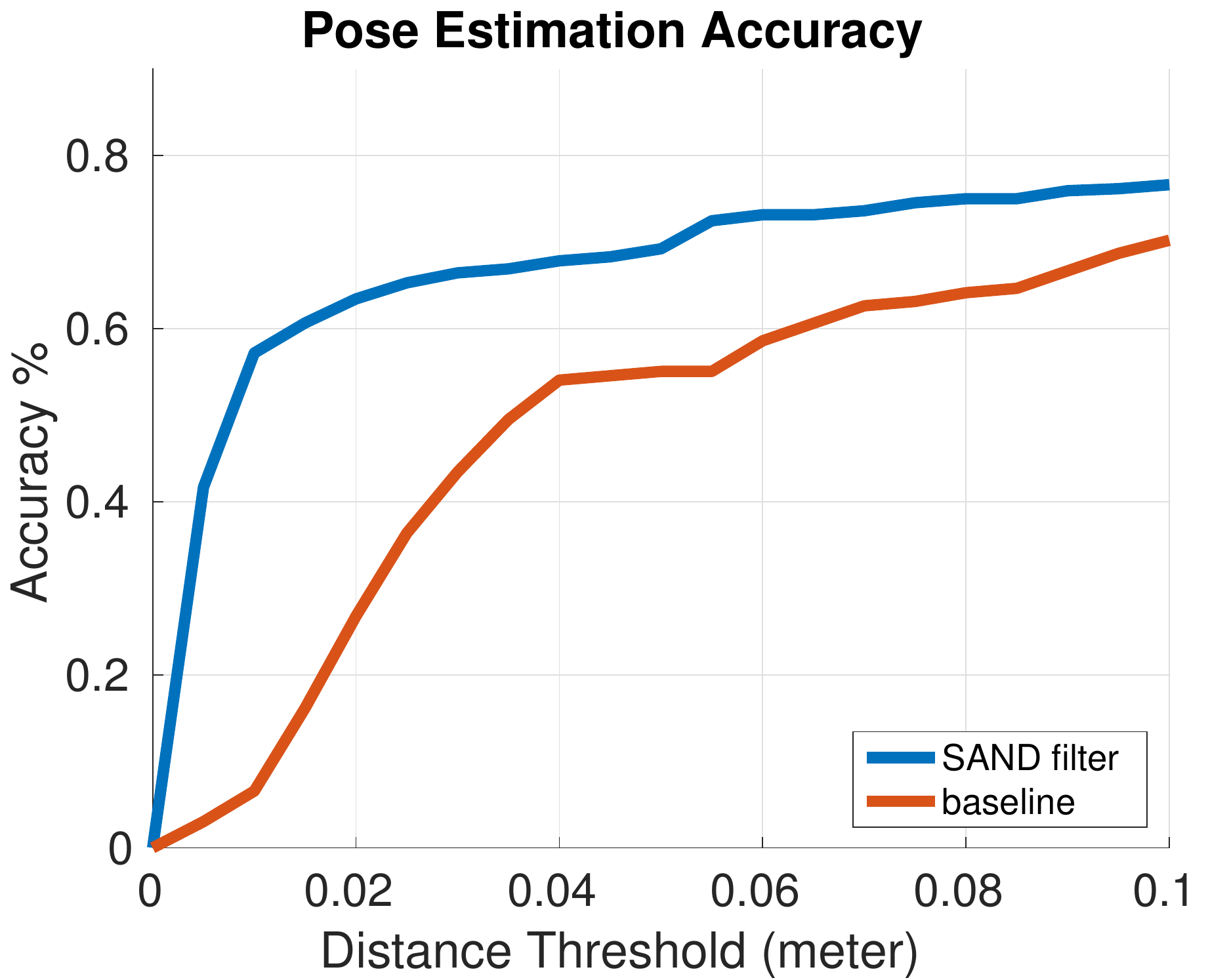}
    \caption{Pose estimation accuracy-threshold curves for our SAND filter approach and the baseline approach. The x-axis is the average distance to the ground truth object and the y-axis is the pose accuracy based on the metric.}
\label{fig:sand_faster_compare}
\end{figure}

We first compare the SAND filter with a baseline method where it takes the state-of-art object detector Faster-RCNN~\cite{faster-rcnn} as the first stage and the widely used pose estimation algorithm ICP~\cite{icp} as the second stage. Shown in Figure~\ref{fig:sand_faster_compare}, the SAND filter method is consistently better than the baseline method over all the thresholds. For the tighter distance threshold (e.g. 0.02 meter), our approach is able to achieve over sixty percent accuracy, which greatly outperforms the baseline method. This is due to the ability that our second stage can iteratively narrow down the search space through sampling-based local search. Note that for the baseline approach, the orientation initialization of the ICP for each object in the scene is set to zero which makes the result bias towards specific object configurations (e.g. objects standing on the table). 

\begin{figure}[t]
    \centering
    \includegraphics[width=\linewidth]{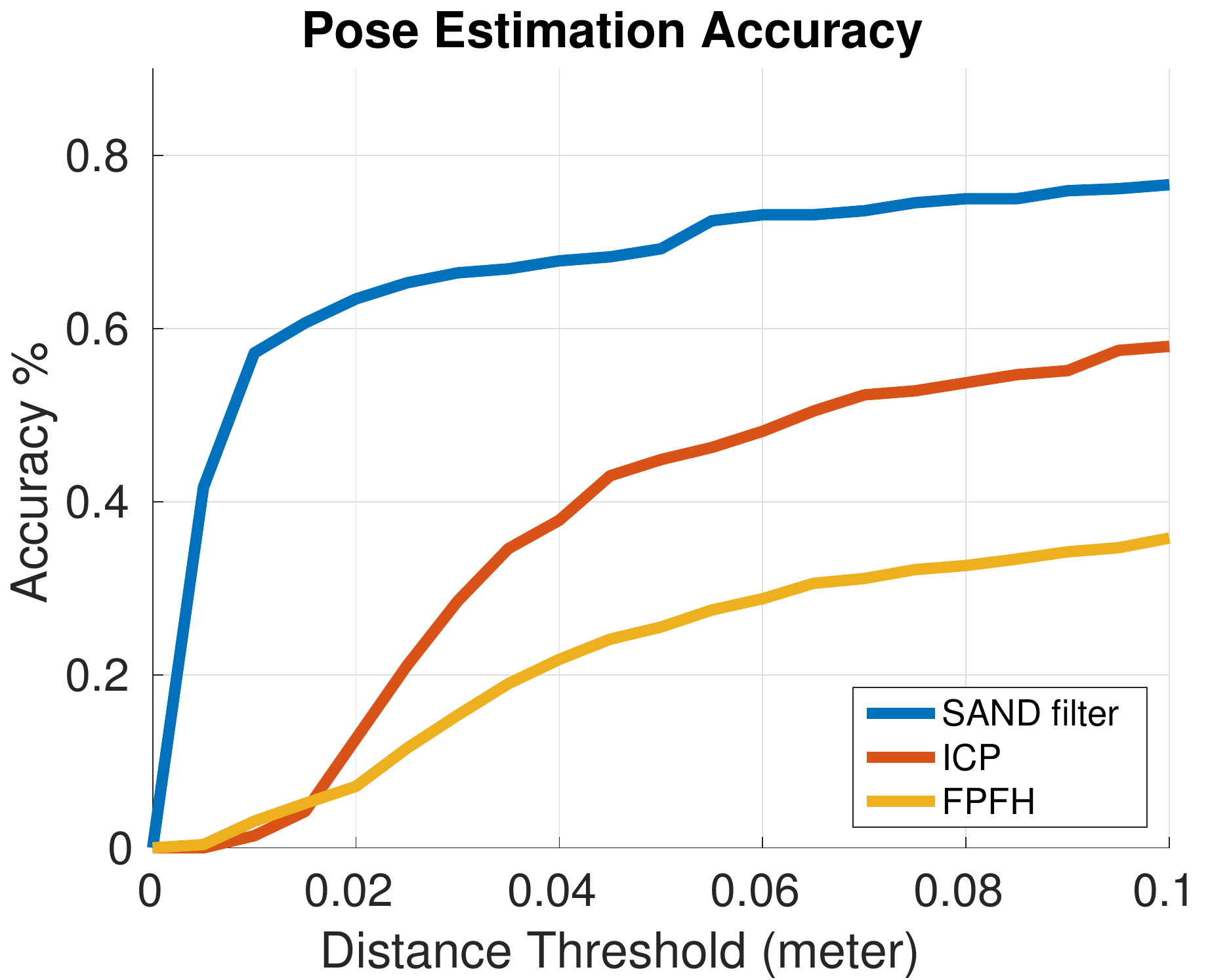}
    \caption{Pose estimation accuracy-threshold curves for comparing different second stage method for the SAND filter approach. The x-axis is the average distance to the ground truth object and the y-axis is the pose accuracy based on the metric.}
\label{fig:sand_icp_compare}
\end{figure}

Then we substitute the second stage in the SAND filter method from generative sampling-based local search method to ICP and Fast Point Feature Histograms (FPFH)~\cite{fpfh}, but remain first stage as the same. The plots in figure~\ref{fig:sand_icp_compare} show that our second stage method is significantly better thant ICP and FPFH with the pyramid of heatmap detections as the first stage.



\subsubsection{Categorical Sorting}
\begin{figure}[t]
\begin{center}
   \includegraphics[width=0.9\linewidth]{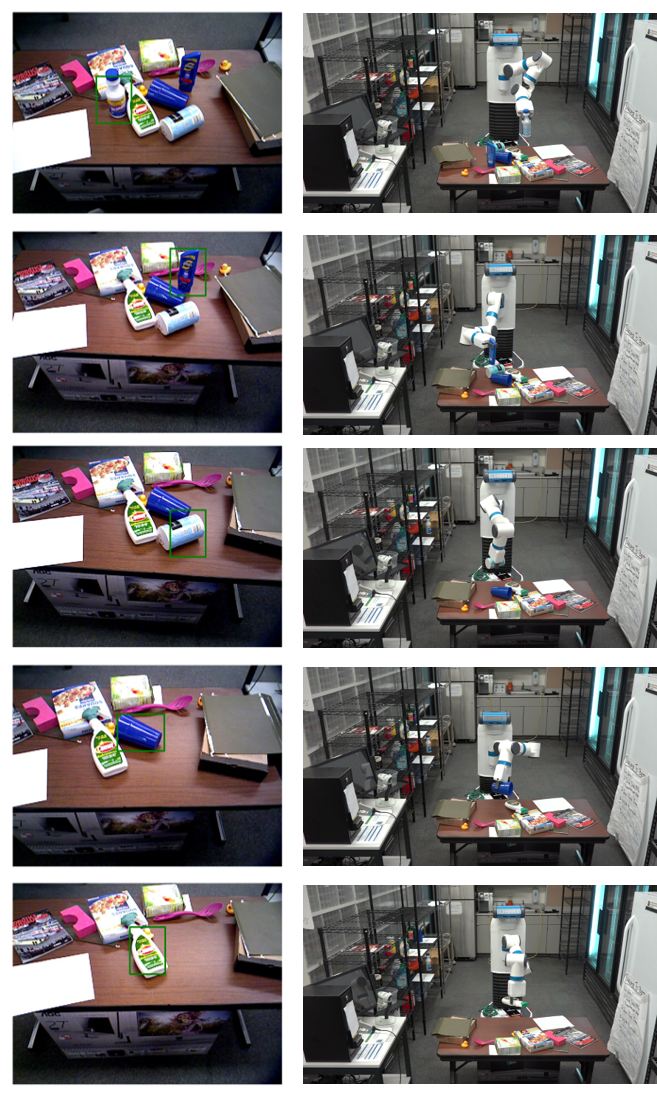}
\end{center}
   \caption{The robot executes categorical sorting tasks based on SAND Filter in one action sequence.}
\label{fig:tasks}
\end{figure}

In the manipulation experiments, the task for the robot is to sort all the objects on a tabletop into different locations on the shelf according to their categories (left column of Figure~\ref{fig:overview}). The robot first performs two-stage method to acquire objects' poses, and further grasp the objects and place them onto the corresponding locations of the shelf one at a time. We divide the 15 objects into three categories: \emph{food \& drink}, \emph{laundry} and \emph{miscellaneous}. The pose estimation will be performed after placing each object. This process is illustrated in Figure~\ref{fig:tasks}.

The robot performs $9$ sorting sequences in total. Because task completion depends on many factors, such as motion planning and grasping, we split the task into three sub-tasks: \emph{pose estimation}, and \emph{grasping}, and \emph{placement}. In this case, we can further analyze the success rate for each sub-task and determine the bottleneck of the entire system.

Table~\ref{tab:manipulation} shows the success rate for each sub-task in our categorical sorting tasks. There are average $4.78$ objects in each task. For pose estimation, if all the detections are lower than a certain threshold, it considers to be a failure. After successful pose estimation, we further inspect object grasping given the matched poses. As for placement task, as long as the robot is able to place the object to its corresponding area on the shelf, we consider it as a success. Since these three sub-tasks are sequential, we consider the entire task is completed if the placement is successful.

\begin{table}[h]
\centering
\resizebox{\columnwidth}{!}{
\begin{tabular}{|c|c|c|c|}
\hline
 & Pose Estimation & Grasping & Task Completion \\ \hline
 Success Rate & 40/42 (0.952) & 37/40 (0.925)  & 37/42 (0.881) \\ \hline
\end{tabular}
}
\caption{The tables shows results of categorical sorting experiments for 9 sequences.}
\label{tab:manipulation}
\end{table}

According to Table~\ref{tab:manipulation}, the main source of failure is grasping since three out of five failed cases are due to grasping. The end-effector of the Fetch robot is a hard gripper and without any tactile sensors. Therefore, the robot essentially performs open-loop grasping on top of the cluttered tabletop without any feedback. For failed pose estimation, our detector fails to locate the object in the scene, which leads to negative pose match. However, considering the challenging nature of our scenes and tasks, $0.881$ total success rate is promising.

\section{Conclusion and Discussion}
\label{sec:conclusion}
In this work, we first present a novel two-stage paradigm, SAND filter, that leverages both CNN and generative sampling-based local search to achieve accurate object detection and six DoF object poses. We further build a manipulation pipeline to perform categorical object sorting tasks.

To perform object manipulation task requires accurate object detection and pose estimation. Our SAND filter enables the robot to perceive the environment regardless of occlusion and clutteredness. We hope our two-stage paradigm would shed light on the challenging nature of perception for manipulation tasks and further progress towards true autonomous manipulation.

In the future, we will enable SAND filter to detect and estimate pose for multiple object instances. Besides, the current SAND filter only works on single observation, and it can be extend to account for sequential observation.


\bibliographystyle{plainnat}
\bibliography{references}

\begin{thebibliography}{32}
\providecommand{\natexlab}[1]{#1}
\providecommand{\url}[1]{\texttt{#1}}
\expandafter\ifx\csname urlstyle\endcsname\relax
  \providecommand{\doi}[1]{doi: #1}\else
  \providecommand{\doi}{doi: \begingroup \urlstyle{rm}\Url}\fi

\bibitem[Besl and McKay(1992)]{icp}
Paul~J Besl and Neil~D McKay.
\newblock Method for registration of 3-d shapes.
\newblock In \emph{Sensor Fusion IV: Control Paradigms and Data Structures},
  volume 1611, pages 586--607. International Society for Optics and Photonics,
  1992.

\bibitem[Ciocarlie et~al.(2014)Ciocarlie, Hsiao, Jones, Chitta, Rusu, and
  {\c{S}}ucan]{ciocarlie2014towards}
Matei Ciocarlie, Kaijen Hsiao, Edward~Gil Jones, Sachin Chitta, Radu~Bogdan
  Rusu, and Ioan~A {\c{S}}ucan.
\newblock Towards reliable grasping and manipulation in household environments.
\newblock In \emph{Experimental Robotics}, pages 241--252. Springer Berlin
  Heidelberg, 2014.

\bibitem[Collet et~al.(2011)Collet, Martinez, and Srinivasa]{collet2011moped}
Alvaro Collet, Manuel Martinez, and Siddhartha~S Srinivasa.
\newblock The moped framework: Object recognition and pose estimation for
  manipulation.
\newblock \emph{The International Journal of Robotics Research}, page
  0278364911401765, 2011.

\bibitem[Dellaert et~al.(1999)Dellaert, Fox, Burgard, and
  Thrun]{dellaert1999monte}
Frank Dellaert, Dieter Fox, Wolfram Burgard, and Sebastian Thrun.
\newblock Monte carlo localization for mobile robots.
\newblock In \emph{Robotics and Automation, 1999. Proceedings. 1999 IEEE
  International Conference on}, volume~2, pages 1322--1328. IEEE, 1999.

\bibitem[Girshick(2015)]{fast-rcnn}
Ross Girshick.
\newblock Fast r-cnn.
\newblock \emph{arXiv preprint arXiv:1504.08083}, 2015.

\bibitem[Girshick et~al.(2014)Girshick, Donahue, Darrell, and Malik]{rcnn}
Ross Girshick, Jeff Donahue, Trevor Darrell, and Jagannath Malik.
\newblock Rich feature hierarchies for accurate object detection and semantic
  segmentation.
\newblock In \emph{Computer Vision and Pattern Recognition (CVPR), 2014 IEEE
  Conference on}, pages 580--587. IEEE, 2014.

\bibitem[Gordon et~al.(1993)Gordon, Salmond, and Smith]{gordon1993novel}
Neil~J Gordon, David~J Salmond, and Adrian~FM Smith.
\newblock Novel approach to nonlinear/non-gaussian bayesian state estimation.
\newblock In \emph{IEE Proceedings F (Radar and Signal Processing)}, volume
  140, pages 107--113. IET, 1993.

\bibitem[Gualtieri et~al.(2017)Gualtieri, ten Pas, and Jr.]{gualtieriPP17}
Marcus Gualtieri, Andreas ten Pas, and Robert~Platt Jr.
\newblock Category level pick and place using deep reinforcement learning.
\newblock \emph{CoRR}, abs/1707.05615, 2017.
\newblock URL \url{http://arxiv.org/abs/1707.05615}.

\bibitem[Hastings(1970)]{Hastings_MCMC}
W.~K. Hastings.
\newblock Monte carlo sampling methods using markov chains and their
  applications.
\newblock \emph{Biometrika}, 57\penalty0 (1):\penalty0 97--109, 1970.
\newblock ISSN 00063444.

\bibitem[He et~al.(2014)He, Zhang, Ren, and Sun]{sppnet}
Kaiming He, Xiangyu Zhang, Shaoqing Ren, and Jian Sun.
\newblock Spatial pyramid pooling in deep convolutional networks for visual
  recognition.
\newblock In \emph{Computer Vision--ECCV 2014}, pages 346--361. Springer, 2014.

\bibitem[Hinterstoisser et~al.(2012)Hinterstoisser, Lepetit, Ilic, Holzer,
  Bradski, Konolige, , and Navab]{hinterstoisser2012accv}
S.~Hinterstoisser, V.~Lepetit, S.~Ilic, S.~Holzer, G.~Bradski, K.~Konolige, ,
  and N.~Navab.
\newblock Model based training, detection and pose estimation of texture-less
  3d objects in heavily cluttered scenes.
\newblock 2012.

\bibitem[Huang et~al.(2015)Huang, Yang, Deng, and Yu]{densebox}
Lichao Huang, Yi~Yang, Yafeng Deng, and Yinan Yu.
\newblock Densebox: Unifying landmark localization with end to end object
  detection.
\newblock \emph{arXiv preprint arXiv:1509.04874}, 2015.

\bibitem[Isard and Blake(1996)]{isard1996contour}
Michael Isard and Andrew Blake.
\newblock Contour tracking by stochastic propagation of conditional density.
\newblock In \emph{European conference on computer vision}, pages 343--356.
  Springer, 1996.

\bibitem[Krizhevsky et~al.(2012)Krizhevsky, Sutskever, and Hinton]{alexnet}
Alex Krizhevsky, Ilya Sutskever, and Geoffrey~E Hinton.
\newblock Imagenet classification with deep convolutional neural networks.
\newblock In \emph{Advances in neural information processing systems}, pages
  1097--1105, 2012.

\bibitem[Long et~al.(2014)Long, Shelhamer, and Darrell]{fcn}
Jonathan Long, Evan Shelhamer, and Trevor Darrell.
\newblock Fully convolutional networks for semantic segmentation.
\newblock \emph{arXiv preprint arXiv:1411.4038}, 2014.

\bibitem[Narayanan and Likhachev(2016)]{Narayanan-RSS-16}
Venkatraman Narayanan and Maxim Likhachev.
\newblock Discriminatively-guided deliberative perception for pose estimation
  of multiple 3d object instances.
\newblock In \emph{Proceedings of Robotics: Science and Systems}, AnnArbor,
  Michigan, June 2016.
\newblock \doi{10.15607/RSS.2016.XII.023}.

\bibitem[Papazov et~al.(2012)Papazov, Haddadin, Parusel, Krieger, and
  Burschka]{papazov2012rigid}
Chavdar Papazov, Sami Haddadin, Sven Parusel, Kai Krieger, and Darius Burschka.
\newblock Rigid 3d geometry matching for grasping of known objects in cluttered
  scenes.
\newblock \emph{The International Journal of Robotics Research}, page
  0278364911436019, 2012.

\bibitem[Redmon et~al.(2015)Redmon, Divvala, Girshick, and Farhadi]{yolo}
Joseph Redmon, Santosh Divvala, Ross Girshick, and Ali Farhadi.
\newblock You only look once: Unified, real-time object detection.
\newblock \emph{arXiv preprint arXiv:1506.02640}, 2015.

\bibitem[Ren et~al.(2015)Ren, He, Girshick, and Sun]{faster-rcnn}
Shaoqing Ren, Kaiming He, Ross Girshick, and Jian Sun.
\newblock Faster r-cnn: Towards real-time object detection with region proposal
  networks.
\newblock In \emph{Advances in Neural Information Processing Systems}, pages
  91--99, 2015.

\bibitem[Rusu(2009)]{fpfh}
Radu~Bogdan Rusu.
\newblock \emph{Semantic 3D Object Maps for Everyday Manipulation in Human
  Living Environments}.
\newblock PhD thesis, Computer Science department, Technische Universitaet
  Muenchen, Germany, October 2009.

\bibitem[Sermanet et~al.(2013)Sermanet, Eigen, Zhang, Mathieu, Fergus, and
  LeCun]{overfeat}
Pierre Sermanet, David Eigen, Xiang Zhang, Micha{\"e}l Mathieu, Rob Fergus, and
  Yann LeCun.
\newblock Overfeat: Integrated recognition, localization and detection using
  convolutional networks.
\newblock \emph{arXiv preprint arXiv:1312.6229}, 2013.

\bibitem[Simonyan and Zisserman(2014)]{vgg}
Karen Simonyan and Andrew Zisserman.
\newblock Very deep convolutional networks for large-scale image recognition.
\newblock \emph{arXiv preprint arXiv:1409.1556}, 2014.

\bibitem[Sui et~al.(2015)Sui, Jenkins, and Desingh]{sui2015iros}
Zhiqiang Sui, Odest~Chadwicke Jenkins, and Karthik Desingh.
\newblock Axiomatic particle filtering for goal-directed robotic manipulation.
\newblock In \emph{Intelligent Robots and Systems (IROS), 2015 IEEE/RSJ
  International Conference on}, pages 4429--4436. IEEE, 2015.

\bibitem[Sui et~al.(2017{\natexlab{a}})Sui, Xiang, Jenkins, and
  Desingh]{sui2017ijrr}
Zhiqiang Sui, Lingzhu Xiang, Odest~C Jenkins, and Karthik Desingh.
\newblock Goal-directed robot manipulation through axiomatic scene estimation.
\newblock \emph{The International Journal of Robotics Research}, 36\penalty0
  (1):\penalty0 86--104, 2017{\natexlab{a}}.

\bibitem[Sui et~al.(2017{\natexlab{b}})Sui, Zhou, Zeng, and
  Jenkins]{sui2017iros}
Zhiqiang Sui, Zheming Zhou, Zhen Zeng, and Odest~Chadwicke Jenkins.
\newblock Sum: Sequential scene understanding and manipulation.
\newblock In \emph{2017 IEEE/RSJ International Conference on Intelligent Robots
  and Systems (IROS)}, pages 3281--3288, Sept 2017{\natexlab{b}}.
\newblock \doi{10.1109/IROS.2017.8206164}.

\bibitem[Ten~Pas and Platt(2016)]{ten2016localizing}
Andreas Ten~Pas and Robert Platt.
\newblock Localizing handle-like grasp affordances in 3d point clouds.
\newblock In \emph{Experimental Robotics}, pages 623--638. Springer, 2016.

\bibitem[Uijlings et~al.(2013)Uijlings, van~de Sande, Gevers, and
  Smeulders]{ss}
Jasper~RR Uijlings, Koen~EA van~de Sande, Theo Gevers, and Arnold~WM Smeulders.
\newblock Selective search for object recognition.
\newblock \emph{International journal of computer vision}, 104\penalty0
  (2):\penalty0 154--171, 2013.

\bibitem[Vapnik et~al.(1997)Vapnik, Golowich, and Smola]{vapnik97nips}
V.~Vapnik, S.E. Golowich, and A.~Smola.
\newblock Support vector method for function approximation, regression
  estimation, and signal processing.
\newblock 1997.

\bibitem[Varley et~al.(2015)Varley, Weisz, Weiss, and Allen]{Varley2015IROS}
J.~Varley, J.~Weisz, J.~Weiss, and P.~Allen.
\newblock Generating multi-fingered robotic grasps via deep learning.
\newblock In \emph{2015 IEEE/RSJ International Conference on Intelligent Robots
  and Systems (IROS)}, pages 4415--4420, Sept 2015.
\newblock \doi{10.1109/IROS.2015.7354004}.

\bibitem[Xiang et~al.(2017)Xiang, Schmidt, Narayanan, and
  Fox]{xiang2017posecnn}
Yu~Xiang, Tanner Schmidt, Venkatraman Narayanan, and Dieter Fox.
\newblock Posecnn: A convolutional neural network for 6d object pose estimation
  in cluttered scenes.
\newblock \emph{arXiv preprint arXiv:1711.00199}, 2017.

\bibitem[Zeiler and Fergus(2014)]{zeiler2014visualizing}
Matthew~D Zeiler and Rob Fergus.
\newblock Visualizing and understanding convolutional networks.
\newblock In \emph{European conference on computer vision}, pages 818--833.
  Springer, 2014.

\bibitem[Zitnick and Doll{\'a}r(2014)]{edgebox}
C~Lawrence Zitnick and Piotr Doll{\'a}r.
\newblock Edge boxes: Locating object proposals from edges.
\newblock In \emph{Computer Vision--ECCV 2014}, pages 391--405. Springer, 2014.

\end{thebibliography}

\end{document}